\documentclass[journal]{IEEEtranTIE}
\usepackage{graphicx}
\usepackage{cite}
\usepackage{picinpar}
\usepackage{amsmath}
\usepackage{url}
\usepackage{flushend}
\usepackage[latin1]{inputenc}
\usepackage{colortbl}
\usepackage{soul}
\usepackage{multirow}
\usepackage{pifont}
\usepackage{color}
\usepackage{alltt}
\usepackage[hidelinks]{hyperref}
\usepackage{enumerate}
\usepackage{siunitx}
\usepackage{breakurl}
\usepackage{epstopdf}
\usepackage{pbox}
\usepackage{hhline}

\begin{document}
\title{	Flying through cluttered and dynamic environments with LiDAR}

\author{
	\vskip 1em
	
	Huajie Wu,
	Wenyi Liu, Yunfan Ren, Zheng Liu,
        Hairuo Wei,
        Fangcheng Zhu, Haotian Li,  and Fu Zhang

	\thanks{
The authors gratefully acknowledge DJI for fund support
and Livox Technology for equipment support during the project. This work is supported by Hong Kong General Research Fund (GRF) with grant number 17204523.
 
    The authors are with the Mechatronics and Robotic Systems (MaRS) Laboratory, Department of Mechanical Engineering, University of Hong Kong, Hong Kong SAR, China. \textit{(Huajie Wu and Wenyi Liu contributed equally to this work.) (Corresponding author: Fu Zhang.) }
	}
}

\maketitle
	
\begin{abstract}
Navigating unmanned aerial vehicles (UAVs) through cluttered and dynamic environments remains a significant challenge, particularly when dealing with fast-moving or sudden-appearing obstacles. This paper introduces a complete LiDAR-based system designed to enable UAVs to avoid various moving obstacles in complex environments. Benefiting the high computational efficiency of perception and planning, the system can operate in real time using onboard computing resources with low latency. For dynamic environment perception, we have integrated our previous work, M-detector, into the system. M-detector ensures that moving objects of different sizes, colors, and types are reliably detected. For dynamic environment planning, we incorporate dynamic object predictions into the integrated planning and control (IPC) framework, namely DynIPC. This integration allows the UAV to utilize predictions about dynamic obstacles to effectively evade them. We validate our proposed system through both simulations and real-world experiments. In simulation tests, our system outperforms state-of-the-art baselines across several metrics, including success rate, time consumption, average flight time, and maximum velocity. In real-world trials, our system successfully navigates through forests, avoiding moving obstacles along its path.
\end{abstract}


\markboth{IEEE TRANSACTIONS ON INDUSTRIAL ELECTRONICS}%
{}

\definecolor{limegreen}{rgb}{0.2, 0.8, 0.2}
\definecolor{forestgreen}{rgb}{0.13, 0.55, 0.13}
\definecolor{greenhtml}{rgb}{0.0, 0.5, 0.0}

\section{Introduction}

\IEEEPARstart{I}{n} recent years, the development of lightweight and high-precision sensors, such as Light Detection and Ranging sensors (LiDAR), event cameras, and depth cameras, has significantly advanced the autonomous flight capabilities of unmanned aerial vehicles (UAVs) or drones. This technological progress has facilitated the widespread application of drones across various industries, including agricultural spraying \cite{istiak2023adoption}, product delivery \cite{lammers2023airborne}, inspection \cite{liu2024lidar}, and search and rescue \cite{lyu2023unmanned}. These applications have notably enhanced production efficiency, reduced costs, and driven economic growth within these sectors.

\begin{figure}[!t]\centering
	\includegraphics[width=8cm]{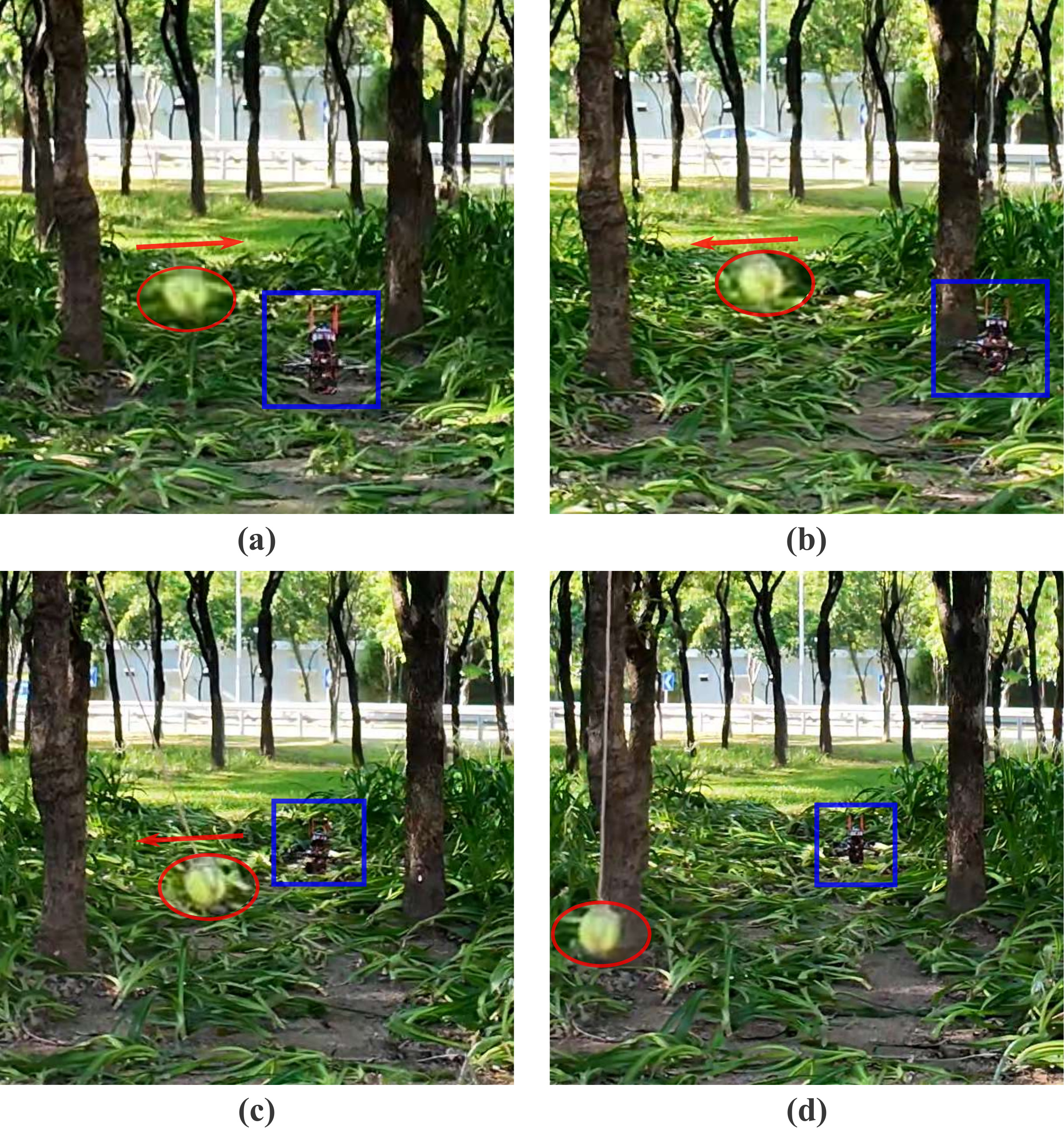}
	\caption{\textbf{UAV Navigating through a Cluttered and Dynamic Environment.}  A pendulum made of a tennis ball is introduced as a moving obstacle along the UAV's path. The red circle highlights the enlarged obstacle, while the red arrow indicates its direction of movement at each time point. The blue box represents the UAV's position. (a) The pendulum swung to the right. The UAV detected a potential collision and planned to move to the right to avoid it. (b) As the pendulum reached its farthest right position and started to move left, the UAV moved to the right to avoid it. (c) With the pendulum now moving left, the UAV successfully navigated past the dynamic obstacle. (d) The UAV had cleared the pendulum and continued its flight through the cluttered environment.}\label{fig:cover}
\end{figure}

However, UAVs often face challenges when navigating environments with unknown and moving obstacles, such as other UAVs, unmanned ground vehicles, animals, pedestrians, and other moving objects. UAVs may crash if they collide with these obstacles, leading to significant economic losses and posing risks to personal safety. Given the ubiquity of dynamic obstacles in real-world settings, it is crucial to equip UAVs with the ability to avoid these obstacles, thereby enhancing their safety and robustness in cluttered and dynamic environments (Fig. \ref{fig:cover}).

Several works \cite{tordesillas2022panther, lu2022perception, chen2023flying, xu2023vision, lin2020robust, falanga2020dynamic, kong2021avoiding, lu2023fapp} have attempted to address this issue. The typical system involves in state estimation, dynamic environment perception, dynamic environment planning, and control. Despite these efforts, several challenges remain when deploying such systems on UAVs, particularly due to their limited onboard computing resources. Key challenges include:

\begin{enumerate}[1)]
	\item The complexity of real-world environments poses significant challenges for UAVs in avoiding obstacles in unknown, cluttered, and dynamic scenes. Existing approaches often can only evade dynamic obstacles in relatively open environments.
 
	\item In real-world scenarios, UAVs must be capable of avoiding fast-moving or sudden-appearing objects. This capability demands a system with low latency from perception to action to ensure timely and effective responses.
 
    \item The variety of dynamic objects in the real world emphasizes the necessity of developing a system that can effectively handle diverse objects, irrespective of their sizes, classes, or colors.
\end{enumerate}

In this paper, we propose a general framework that enables safe and real-time flight in cluttered and dynamic environments with a LiDAR sensor. We address the aforementioned challenges by incorporating our previous detection work, M-detector \cite{wu2024moving}, in the detection module and introducing a new integrated planning and control framework for dynamic objects (DynIPC), which is based on our previous work IPC \cite{liu2023integrated}. Specifically, our main contributions are:

\begin{enumerate}[1)]
	\item  We present a complete LiDAR-based system that enables UAVs to navigate through unknown, cluttered, and dynamic environments.
	\item The low latency of M-detector and DynIPC allows the system to successfully avoid fast-moving or sudden-appearing objects with a latency of around 11 ms.
        \item The high generality of M-detector enables the system to perceive and avoid various dynamic objects, regardless of their sizes, classes, or colors, significantly enhancing the UAV's autonomous capabilities in complex and unknown scenes. 
        
\end{enumerate}

The rest of this article is organized as follows: Section \ref{sec:related} discusses relevant research works. An overview of the complete framework and the details of each key module are presented in Sections \ref{sec:framework}, \ref{sec:perception}, \ref{sec:map}, and \ref{sec:planning} respectively. The benchmarks in simulation and real-world experiment validations are reported in Section \ref{sec:result}. Finally, Section \ref{sec:conclusion} concludes the article.

\begin{figure*}[htb]\centering
	\includegraphics[width=0.8\linewidth]{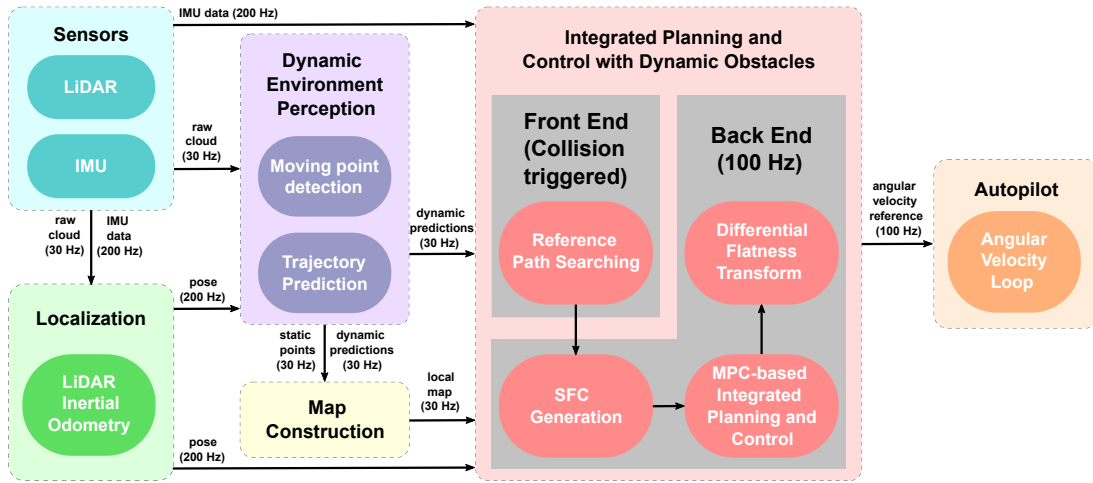}
	\caption{The overview of our system.}\label{fig:framework}
\end{figure*}

\section{Related Work}
\label{sec:related}
\subsection{Dynamic environment perception}
For the perception of static and dynamic environments, researchers typically utilize a variety of sensors, including depth cameras \cite{tordesillas2022panther, lu2022perception, chen2023flying, xu2023vision, lin2020robust}, event cameras \cite{falanga2020dynamic}, LiDARs \cite{kong2021avoiding, liu2023integrated, lu2023fapp, wu2024moving}, or combinations of these technologies \cite{he2021fast}. 

Depth cameras are a popular choice for UAV dynamic environment perception due to their low cost and lightweight nature. Common algorithms for moving object detection with depth cameras involve clustering points, tracking clusters using Kalman filters, and filtering dynamic objects based on velocity \cite{tordesillas2022panther, lin2020robust}. However, these methods struggle in complex environments where static and dynamic obstacles are interconnected, such as pedestrians walking on the ground. Additionally, in \cite{xu2023vision}, u-depth map is generated and utilized to extract dynamic objects. However, this method considers points with the same depth to be the same object, resulting in a significant loss of accuracy. Learning-based detection networks have also been explored \cite{chen2023flying, lu2022perception}, but they are limited to segmenting only pre-trained objects regardless of their movement status and lack generalizability to other dynamic objects. Furthermore, the sensing range of depth cameras, typically within 10 meters, is insufficient for UAVs in challenging scenes.

Event cameras, known for their natural sensitivity to dynamic objects, have also been studied for dynamic perception. Some research has equipped UAVs with event cameras to avoid dynamic objects, achieving latencies as low as 3.5 milliseconds \cite{falanga2020dynamic}. However, event cameras alone cannot fully perceive static environments, making them inadequate for navigating UAVs through cluttered and dynamic environments that contain many static obstacles. While integrating event cameras with other sensors, such as depth cameras, can mitigate this issue \cite{he2021fast}, it introduces new challenges in multi-sensor calibration. Additionally, the increased payload for UAVs and the high cost of event cameras impede their widespread commercial application.

With the commercialization of LiDAR, researchers are increasingly equipping UAVs with this sensor to avoid dynamic objects \cite{kong2021avoiding, lu2023fapp, liu2023integrated, wu2024moving} due to its affordable price, long sensing range, high-frequency scanning, and precise depth measurements. Most of these studies \cite{kong2021avoiding, liu2023integrated} employed LiDARs without distinguishing between dynamic and static points, simply relying on high-frequency updates of scanning data to the map for obstacle avoidance. These reactive methods depend heavily on accumulated points and fail to predict and avoid fast-moving objects in advance due to the lack of trajectory prediction. Other approaches  \cite{lu2023fapp} classify dynamic objects based on clustering results, but these methods face the similar problem with the clustering methods used for depth cameras. Besides, with the point numbers increasing, the time consumption of this method will significantly increase. In our previous work, M-detector \cite{wu2024moving}, we demonstrated the potential of using LiDAR for UAV dynamic obstacle avoidance. M-detector's features of low latency, high generalizability, and high accuracy enable UAVs to perceive dynamic environments instantly and accurately. However, a comprehensive system for UAV navigation in cluttered and dynamic environments has not been thoroughly studied.

In this work, we utilize LiDAR for environmental perception and deploy M-detector to classify dynamic objects. This approach allows the UAV to detect various dynamic objects regardless of their sizes, classes, or colors while preserving complete information about the static environment. Additionally, the low latency of M-detector provides sufficient time for UAVs to respond to fast-moving or sudden-appearing objects, enhancing UAVs' safety and robustness.

\subsection{Dynamic environment planning}
The algorithm for UAV planning in static environments has been extensively studied over the past decades. However, planning algorithms for UAVs in dynamic environments have not received sufficient attention. Recently, several works have attempted to integrate dynamic object information into classic planning algorithms to enable UAVs to avoid dynamic obstacles. For instance, in \cite{falanga2020dynamic}, artificial potential field (APF) is used to avoid dynamic obstacles by generating a potential field that repels the UAV away from detected obstacles. While this method achieves low latency, it is easy to get trapped in local minima. Besides, if there is a narrow passage, the repulsive forces from obstacles on both sides can overpower the attractive force towards the goal, making it difficult or impossible for the UAV to pass through it. Overall, this method can not work well in complicated environments. 

Optimization-based methods have also been explored for dynamic environment planning \cite{xu2023vision, tordesillas2022panther, chen2023flying, lu2022perception, lu2023fapp}. These methods allow the optimization function to be customized according to different tasks and constraints, making them suitable for specific problems. However, the solution process heavily depends on the initial value and may fall into local minima. Additionally, if the optimization problem is overly complex, the time required for computation can increase significantly.

Besides, model predictive control (MPC) has gained increasing attention for avoiding dynamic obstacles in recent years \cite{lin2020robust, liu2023integrated, xu2022dpmpc}. However, current methods either formulate dynamic objects as non-linear constraints \cite{lin2020robust}, resulting in heavy computation loads, or only consider the nearest dynamic obstacle, making it challenging to handle complex scenes with multiple dynamic objects \cite{xu2022dpmpc}. In our previous work \cite{liu2023integrated}, we did not specifically consider the dynamic object information, formulating the problem as a Quadratic Problem (QP) that can be solved at high frequencies (i.e., 100 hz). While this approach allows for reactive avoidance of dynamic objects, it falls short when dealing with fast-moving objects due to the lack of trajectory prediction for moving obstacles.

In this work, we incorporate dynamic obstacle predictions into the Integrated Planning and Control (IPC) framework to create a new method called DynIPC. This method retains IPC's original high-frequency characteristics while also considering the predicted information of dynamic objects, enabling UAVs to avoid them in advance. This integration ensures that UAVs can navigate cluttered and dynamic environments more effectively and safely.

\section{problem statement and proposed framework}
\label{sec:framework}
\subsection{Problem Statement}
In this work, we address a tough challenge for UAVs: navigating through environments that contain both static and various dynamic objects, such as thrown balls, pedestrians, or other agents. The static obstacles are cluttered and complex, while the dynamic objects may be of fast velocities, different sizes, and varying geometries. The UAV must successfully navigate through this challenging environment and reach its target safely, with all processes executed using the onboard computer.

\subsection{Dynamic Avoidance Framework}
We have adopted a mainstream framework for dynamic obstacle avoidance, which includes modules for localization, dynamic environment perception, map construction, and integrated planning and control with dynamic objects. The specified workflow is illustrated in Fig. \ref{fig:framework}.

Utilizing the sensor information, i.e. point cloud and IMU data, LiDAR Inertial Odometry (LIO) is performed at first to determine the current pose of the UAV. The calculated pose, along with the point cloud, is then used in the dynamic environment perception module to distinguish moving and static points and to predict the trajectories of dynamic objects (see section \ref{sec:perception}). Subsequently, both the static points and the predicted positions of dynamic objects are used to construct a map (see section \ref{sec:map}). Finally, an integrated planning and control algorithm is designed to generate commands for the UAV to maneuver safely, avoiding both static and dynamic obstacles present in the map (see section \ref{sec:planning}).

\section{dynamic environment perception}
\label{sec:perception}
\subsection{Moving Point Detection}
To enable the UAV to fly through the cluttered and dynamic environments, detecting the moving objects accurately with a low latency is the vitally important part. To reduce detection latency while ensuring that various objects are detected successfully, we employ our previously developed method, M-detector \cite{wu2024moving}, to classify points as static or dynamic. M-detector leverages the current point cloud and pose to instantly classify the state of the points, without relying on future point clouds or a prior map. This makes it an ideal detection method for onboard dynamic detection in unknown environments. M-detector operates in two modes: point-out mode and frame-out mode. The point-out mode can detect the moving point after its arrival instantly. While the frame-out mode detects the moving point after the steps of clustering and region growth. For this work, we adopt the frame-out mode, which provides more accurate detection at the cost of affordable latency (typically around ten milliseconds). This balance between accuracy and latency makes frame-out mode highly suitable for our UAV's dynamic object detection needs.

\subsection{Trajectory Prediction}
After detecting the moving points, tracking dynamic objects and predicting their future trajectories are crucial for effective obstacle avoidance in UAV navigation. We first cluster the dynamic points classified by M-detector using the DBSCAN algorithm \cite{ester1996density}. The dynamic objects are modeled as cuboids, with the cluster center representing the object's center. The size of the object in each direction $(x, y, z)$ is determined by the farthest distance from the center to the points within the cluster along the corresponding axis. Following this, we employ the Hungarian algorithm \cite{kuhn1955hungarian} for frame-to-frame association of multiple objects. The association metric is based on the errors between predicted and measured positions and sizes, ensuring accurate association of various objects across different frames.

Once several historical positions of an object are obtained, we employ polynomial fitting to estimate its future trajectories. This method can converge quickly, making it beneficial for predicting the trajectories of fast-moving objects. Given the fact that we only fit the objects' trajectory in a short period (within 2 seconds), we consider two motion models, constant velocity and free fall, to predict the obstacles' future positions. These correspond to a linear curve and a quadratic curve with the quadratic coefficient close to half the gravitational acceleration, respectively. Specifically, we initially fit the trajectory using a quadratic curve. If the quadratic coefficient approximates half the gravitational acceleration, we classify the movement as free fall and select the quadratic curve as the trajectory. Otherwise, we classify the movement as constant velocity and perform linear fitting. Based on the fitted trajectory curves, we can infer the future positions of the obstacles.

During flight, new obstacles may appear, or historical obstacles may no longer be detected. For new obstacles, we initially use their detected positions for association. After accumulating at least three detections, polynomial fitting is executed to estimate the predicted positions. For historical objects without new detections, we reserve them for a brief period. If there is no detection within a certain period, such as 1.0 second, we remove the corresponding trajectory curve from the map.

\section{Map construction}
\label{sec:map}
After obtaining the static points and dynamic objects' sizes and trajectory predictions, it is important to represent this information effectively for subsequent modules. ROG-Map \cite{ren2023rog}, the same map structure detailed in IPC, is employed in our system. In dynamic scenes, the original ROG-Map can cause ghosting due to the points of moving objects remaining in the map. To address this, our approach enhances the original ROG-Map by incorporating two layers: the static layer and the dynamic layer, to better represent static obstacles and dynamic objects' trajectories.

The static layer is constructed and updated in the same manner as the original ROG-Map but with one key difference: only static points are used for its construction, rather than the entire raw point cloud as in the original ROG-Map. This prevents the influence of dynamic points in the map and thus ensures a more accurate representation of the static environment. For the dynamic layer, we calculate future positions of dynamic objects based on the predicted trajectories from the perception module. Corresponding grids and their neighbors, within the sizes of the dynamic objects, are labeled as dynamically occupied. The dynamic layer uses a binary state system: zero for free and one for dynamically occupied. Each time the dynamic object trajectory prediction is updated, the previously labeled dynamically occupied grids are cleared, and the new positions are labeled accordingly. 

To ensure the UAV's safety, we implement two inflation strategies. For static layer, the same inflation strategy used in IPC, inflating obstacles by the quadrotor's radius $d_{rad}$, is employed. For dynamic layer, obstacles are inflated according to their velocities. The inflation distance is calculated using $d_{inf} \ = \ d_{rad} + k\cdot vel$, where $k$ denotes the weight of velocity (0.01 in our configuration), and $vel$ is the dynamic obstacles' velocity. When querying the grid states, results from both the static and dynamic layers are considered. If either layer indicates an occupied state, the grid is deemed occupied.

\begin{figure}[!t]\centering
	\includegraphics[width=8cm]{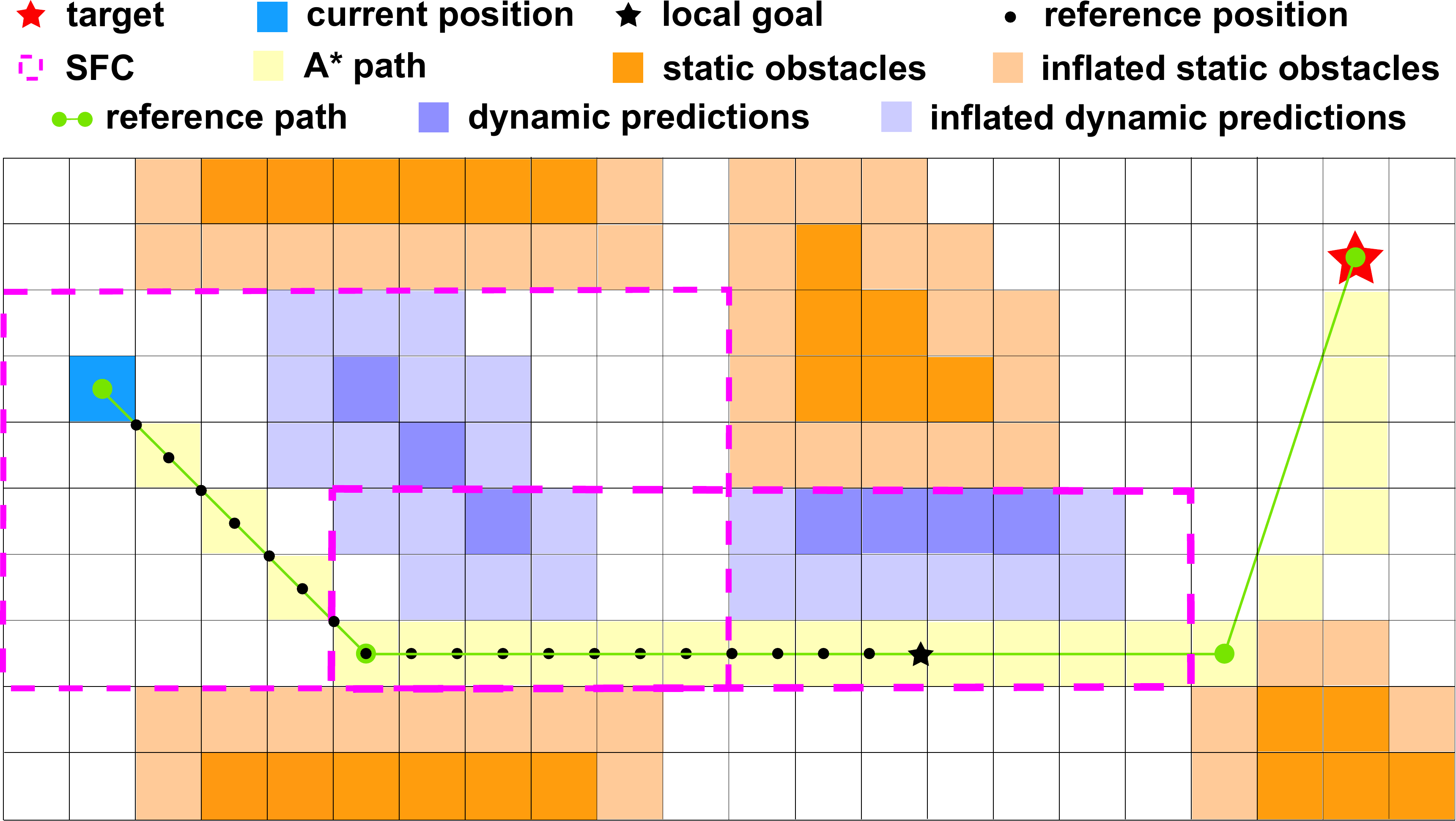}
	\caption{Reference path searching based on the local map in a simplified 2D case.}\label{fig:referencepath}
\end{figure}

\section{integrated planning and control with dynamic objects}
\label{sec:planning}

After obtaining the map containing both static points and dynamic object predictions, our system generates control actions directly in a short time (within 10 ms) to avoid static and dynamic obstacles successfully. This process involves four sequential steps: reference path search, safe flight corridor (SFC) generation, MPC-based planning and control, and differential flatness transform. The differential flatness transform is identical to that used in IPC and will not be detailed here. Instead, we will focus on the remaining three steps and the replanning strategy.

\subsection{Reference Path Search}
We improved the A* algorithm \cite{hart1968formal} to search for a collision-free and dynamic-aware path that avoids both static and dynamic obstacles based on the latest map data. This step ensures that there is a reference path that is free from both static and dynamic obstacles and effectively guides the UAV to the target. Unlike the planning strategy in IPC, our new strategy additionally considers dynamic obstacle predictions (the purple grids in Fig. \ref{fig:referencepath}). As illustrated in Fig. \ref{fig:referencepath}, only grids that are both dynamically and statically safe are regarded as passable in the A* algorithm. This improvement ensures that the path is far away from dynamic objects, enhancing the UAV's ability to navigate through complex environments. After determining a safe path using the A* algorithm, the same pruning strategy used in IPC is adopted to obtain a piecewise short path. This refined path serves as the reference path for the subsequent planning and control steps.

\subsection{SFC Generation}
\label{subsec:sfc}
Next, we generate several safe flight corridors (SFC) around the UAV's current position and along the reference path. The SFC provides the flexible, static-obstacle-free region through which the UAV can navigate, enhancing the flight safety. While we retain the same SFC generation algorithm used in IPC, our approach differs in a key aspect: we use only static points to generate the SFC, rather than all points as in IPC. By focusing solely on static points for SFC generation, we effectively enlarge the safe space for the UAV. This approach not only maximizes the available navigable area but also reduces the likelihood of corridor generation failure in environments with numerous dynamic objects.

The generated SFCs serve as hard constraints in subsequent steps to ensure the UAV does not collide with the static environment. While predictions of dynamic obstacles are incorporated into the objective function as soft constraints, guiding the UAV to avoid dynamic obstacles, as introduced
in subsection \ref{subsec:mpc}.

\subsection{MPC-based Integrated Planning and Control}
\label{subsec:mpc}
Following the generation of the Safe Flight Corridor (SFC), we integrate planning and control into one framework using Model Predictive Control (MPC). This module optimizes the control actions for the UAV within the SFC, ensuring it maintains a safe distance from both static and dynamic obstacles.


In this step, the input of MPC includes the reference positions (obtained from the reference path as in IPC), dynamic object predictions (calculated in Section \ref{sec:perception}), the safe flight corridors (generated in Subsection \ref{subsec:sfc}), the UAV's current state (provided by the LIO module), and all necessary model and kinodynamic constraints. The output is optimal control actions.

We discretize this problem with a time step $\Delta t$ and formulate it with $N$ steps, defining the MPC horizon as $\tau = N\Delta t$. At each step, there are $K_n$ dynamic obstacles predicted, and then the corresponding optimization problem is designed as follows: 
\begin{small}
\begin{subequations}
    \begin{align}
        \begin{split}
            \underset{\mathbf{u}_k}{{\min}} \quad &\sum_{n=1}^N({\left\|\mathbf{p}_{ref,n}-\mathbf{p}_n\right\|}_{\mathbf{R}_p}^2 + {\left\|\mathbf{u}_{n-1}\right\|}_{\mathbf{R}_u}^2 ) \\ 
            &+ \sum_{n=0}^{N-2}{\left\|\mathbf{u}_{n+1}-\mathbf{u}_{n}\right\|}_{\mathbf{R}_c}^2 + {\left\|\mathbf{v}_N\right\|}_{\mathbf{R}_{v,N}}^2 + {\left\|\mathbf{a}_N\right\|}_{\mathbf{R}_{a,N}}^2\\
            &+ \sum_{n=0}^{N}\sum_{k=0}^{K_n}\max(d_{dyn}^2 - {\left\|\mathbf {p}_{n}-\mathbf {p}_{obs, n, k}\right\|}{_{\mathbf {R}_o}^2},\ 0) \\
              \label{eq:cost}
        \end{split} \\
        \mathbf{s.t.} \quad & \mathbf{x}_{n} = \mathbf f_d(\mathbf{x}_{n-1}, \mathbf{u}_{n-1}), \quad n = 1, 2, \cdots, N \label{eq:st_model} \\
        & \mathbf{x}_0 = [\mathbf{p}_{odom}, \mathbf{v}_{odom}, \mathbf{a}_{odom}]^T \label{eq:init} \\
        & |v_{i,n}| \leq |v_{i,max}|, |j_{i,n}| \leq |j_{i,max}|, i = x,y,z \label{eq:st1} \\
        & |a_{j,n}| \leq |a_{j,max}|, j = x,y \label{eq:st2} \\
        & a_{z,min} \leq a_{z,n} \leq a_{z,max} \label{eq:st3} \\
        & \mathbf{C}_{n} \cdot \mathbf{p}_n - \mathbf{d}_{n} \leq 0 \label{eq:st_pos}
    \end{align}
    \label{eq:mpc}
\end{subequations}
\end{small}
In cost function (\ref{eq:cost}), ${\left\|\mathbf{p}_{ref,n}-\mathbf{p}_n\right\|}_{\mathbf{R}_p}^2$ penalizes the error between the UAV's positions $\mathbf{p}_{n}$ and the reference positions $\mathbf{p}_{ref,n}$, ${\left\|\mathbf{u}_{n-1}\right\|}_{\mathbf{R}_u}^2$ represents the control efforts required, ${\left\|\mathbf{u}_{n+1}-\mathbf{u}_{n}\right\|}_{\mathbf{R}_c}^2$ is included to ensure consistency in control efforts, ${\left\|\mathbf{v}_N\right\|}_{\mathbf{R}_{v,N}}^2$ denotes the terminal velocity, and ${\left\|\mathbf{a}_N\right\|}_{\mathbf{R}_{a,N}}^2$ denotes the terminal acceleration. Besides, the term $\max(d_{dyn}^2 - {\left\|\mathbf {p}_{n}-\mathbf {p}_{obs, n, k}\right\|}{_{\mathbf {R}_o}^2},\ 0)$ is added to maintain a safe distance between the UAV and dynamic obstacles, where $d_{dyn}$ is the minimum safe distance with dynamic obstacles, and $\mathbf {p}_{obs, n, k}$ is the predicted position of $k$th dynamic obstacles at step $n$. As for $\mathbf{R}_p$, $\mathbf{R}_u$, $\mathbf{R}_c$, $\mathbf{R}_{v, N}$, $\mathbf{R}_{a, N}$, and $\mathbf{R}_o$, they denote the weights of the corresponding terms.

The constraints for the MPC problem include model constraints (\ref{eq:st_model}), kinodynamic constraints (\ref{eq:init}-\ref{eq:st3}), and corridor constraints (\ref{eq:st_pos}), which are consistent with those described in our previous paper IPC \cite{liu2023integrated} and the detailed descriptions can be referred to it.

In the reference path searching step, A* algorithm considered predictions of dynamic objects, including their centers and sizes, to search a safe path. We assume that this path is inherently distant from dynamic objects, and the primary goal of MPC is to closely follow these reference positions. To simplify the dynamic cost in the function (\ref{eq:cost}), at each step $n$, the UAV reference position $\mathbf{p}_{ref,n}$ is treated as the estimated optimal UAV position. This reference position is used to calculate the distance $d_{n, k}$ between the UAV and the $k$th dynamic objects prior to optimization. If $d_{n, k}$ is smaller than the minimum safety distance $d_{dyn}$  (configured as 0.8 m), the corresponding cost is included in the cost function. Besides, to simplify the maximum terms in (\ref{eq:cost}), the cost function is reformulated as follows:
\begin{small}
\begin{align}
\label{eq:cost_new}
    \begin{split}
        \underset{\mathbf{u}_k}{{\min}} \quad &\sum_{n=1}^N({\left\|\mathbf{p}_{ref,n}-\mathbf{p}_n\right\|}_{\mathbf{R}_p}^2 + {\left\|\mathbf{u}_{n-1}\right\|}_{\mathbf{R}_u}^2 ) \\ 
        &+ \sum_{n=0}^{N-2}{\left\|\mathbf{u}_{n+1}-\mathbf{u}_{n}\right\|}_{\mathbf{R}_c}^2 + {\left\|\mathbf{v}_N\right\|}_{\mathbf{R}_{v,N}}^2 + {\left\|\mathbf{a}_N\right\|}_{\mathbf{R}_{a,N}}^2 \\
        &- \sum_{n=0}^{N}\sum_{k=0}^{K_m}{\left\|\mathbf {p}_{n}-\mathbf {p}_{obs, n, k}\right\|}{_{\mathbf {R}_o}^2} 
    \end{split} 
\end{align}
\end{small}
where $K_m$ represents the number of dynamic obstacles that need to be considered at step $N$, with $K_m \leq K_n$. This simplification ensures that the cost function comprises only quadratic and linear terms.

The optimization variables are $\mathbf{U}\,= \,[\mathbf{u}_0, \mathbf{u}_1, \cdots, \mathbf{u}_N]$. The positions $\mathbf{p}_n$ can be transformed linearly into $\mathbf{U}$ based on the MPC model. Consequently, this optimization problem (\ref{eq:cost_new}) consists of a quadratic cost and linear constraints in terms of the optimization variables. The problem can be expressed in the form: $\mathbf{f}({\mathbf{u})\ = \ \frac{1}{2}\mathbf{u}^T\mathbf{H}\mathbf{u} + \mathbf{c}^Tu}$. To make this problem a standard convex quadratic problem (QP) and ensure good solution properties, the Hessian matrix $\mathbf{H}$ must be positive semidefinite. This requirement is fulfilled by setting the weight $\mathbf{R}_o$ smaller than the weight $\mathbf{R}_p$. The optimization problem is solved using OSQP-Eigen, a C++ wrapper for OSQP \cite{stellato2020osqp}.

Once the optimal control actions are obtained from the solution, the local trajectories can be calculated based on these control actions, current odometry, and the model. In cases where the solver fails to find a solution at step $k$, the same strategy as IPC is employed to handle solver failures and ensure the UAV takes reasonable control actions.

\subsection{Replanning Strategy}
Lastly, our system incorporates a replanning strategy for safe flight in unknown and complicated environments. If the UAV encounters unforeseen obstacles or if the predicted paths of dynamic objects change, the system executes the replanning process to ensure continuous safe navigation.

Two types of replanning strategies are included. The first is for the front-end, i.e., reference path searching, which is triggered by collisions. When new point cloud data is obtained or dynamic object prediction positions are updated, a collision check is performed. This check determines if there is a collision between the pre-planned reference path (considering the inflation radius) and any static or dynamic obstacles. If a collision is detected, the replanning process is executed.

The second strategy is for the back-end. The back-end steps are designed to run at a relatively high frequency, specifically 100 Hz, to maintain optimal control performance. This high-frequency replanning ensures that the UAV can adapt quickly and effectively to any changes in the environment, maintaining safe and efficient navigation.

\section{experiments and results}
\label{sec:result}
\subsection{Simulation Experiments}

To validate the high performance of our system, simulation experiments are conducted. We compared our system with two state-of-the-art (SOTA) open-source systems: Panther \cite{tordesillas2022panther} and Chen's work \cite{chen2023flying}. These systems share a similar framework to ours, including dynamic perception, front-end path searching, and back-end trajectory optimization. The main difference lies in the control strategy: both Panther and Chen's work send the optimized trajectory, including position ($\mathbf{p}$), velocity ($\mathbf{v}$), and acceleration ($\mathbf{a}$), to the autopilot for trajectory tracking, whereas our system sends throttle and angular velocity commands directly to the autopilot.

\begin{figure}[!t]\centering
	\includegraphics[width=7.0cm]{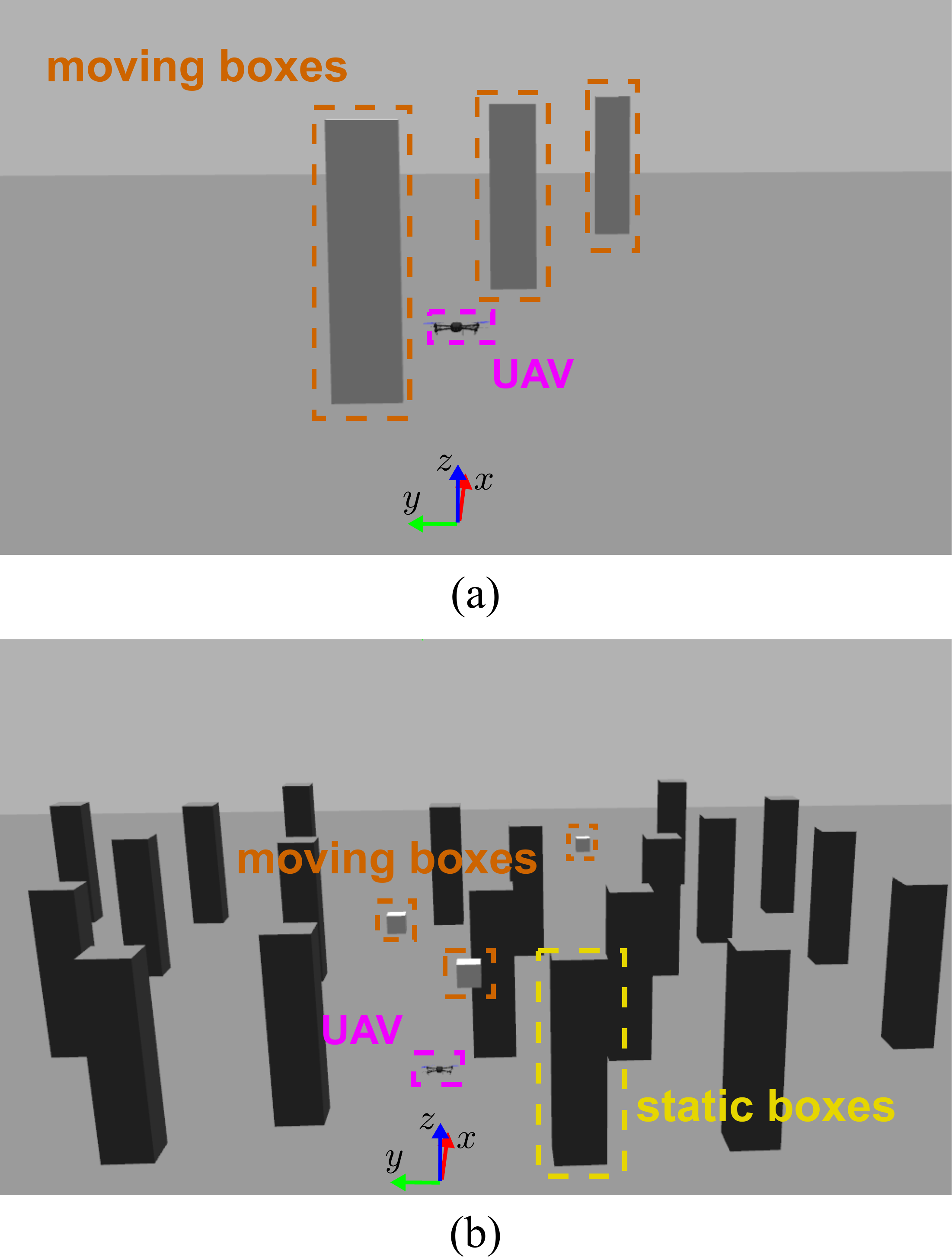}
	\caption{Simulation environments for UAV to fly through the dynamic environments.}\label{fig:simulation}
\end{figure}

All simulation experiments were conducted on a desktop equipped with an Intel i7-1165G7 CPU (2.90 GHz, 8 cores) and 32 GB RAM. We tested the three methods in two different scenarios constructed using GAZEBO. In the first scenario, three dynamic obstacles are in reciprocating motion at a velocity of 6 m/s, as shown in Fig. \ref{fig:simulation}(a). The second scenario is more complicated, featuring numerous static boxes and three dynamic objects in reciprocating motion at a velocity of 2 m/s, as illustrated in Fig. \ref{fig:simulation}(b). 

In both simulation scenarios, the UAV is equipped with a depth camera operating at a frequency of 30 Hz to perceive its surroundings. We chose a depth camera instead of LiDAR because the benchmark methods are designed based on depth cameras. Therefore, our system also uses the point cloud from the depth camera to avoid obstacles, ensuring that performance differences are not due to different sensors. The ground truth for localization is provided for all three methods. Since Chen's work requires a trained network for specific dynamic obstacles, and the obstacles in GAZEBO were not pre-trained, we provided the ground truth for their perception module results, i.e., the static point cloud and the positions and velocities of moving obstacles. Both Panther and our system use the point cloud as input to detect and track dynamic objects. The remaining modules, including planning and control, were kept the same as their open-source versions.

In the simulation experiments, the UAV starts at (0, 0, 2) m, and the target is set as (30, 0, 2) m. The UAV must navigate through the obstacles to reach the goal successfully. Each method is tested 30 times in every scene. We evaluate several performance metrics: the success rate $r$, the mean computation time for dynamic environment perception $T_{per}$, the mean computation time for replanning $T_{replan}$ (including both front-end path searching and back-end trajectory optimization), the average flight time from the start point to the goal $T_{flight}$, and the maximum velocity of the UAV during flight $v_{max}$. The results of these evaluations are presented in Table \ref{tab:comp}.

\begin{table}[!t]
	\renewcommand{\arraystretch}{1.3}
	\caption{The results of Panther, Chen's work, and our system in two simulation scenes.}
	\centering
	\label{tab:comp}
	\resizebox{\columnwidth}{!}{
		\begin{tabular}{l  l l l  l l l}
			\hline\hline \\[-3mm]
                & \multicolumn{3}{c}{Scene1} & \multicolumn{3}{c}{Scene2}\\
                \hhline{~------}
                & Panther & Chen's & Ours & Panther & Chen's & Ours\\ \hline
			$r$ & 0.60 & 0.43 & \textbf{0.97} & 0.07 & 0.90 & \textbf{1.00} \\
  $T_{per}$ (ms) & \textbf{6.55} & -- & 10.63 & \textbf{7.57} & -- & 10.87\\ 
  $T_{replan}$ (ms) & 52.07 & \textbf{2.54} & {3.61} & 41.42 & 4.95 & \textbf{4.68} \\
 $T_{flight}$ (s) & 14.87 & 13.85 & \textbf{10.13} & 17.13 & 15.07 & \textbf{11.02} \\
 $v_{max}$ (m/s) & 2.68 & 2.70 & \textbf{4.87} & 3.89 & 2.64 & \textbf{4.56} \\ [1.4ex]
			\hline\hline
		\end{tabular}
	}
\end{table}

As illustrated in Table \ref{tab:comp}, our system achieved a significantly better success rate in the two scenes. In terms of perception time consumption, our system's time is slightly higher than that of Panther. Chen's method, which utilizes ground truth for perception, lacks effective data for comparison. For replanning time consumption, our system shows a comparable time to Chen's method, while Panther consumes significantly more time. Notably, our system reaches the goal with the minimum flight time (10.13 s) and achieves a maximum velocity of 4.87 m/s, significantly outperforming the other two methods.

In the second scene, Panther nearly fails to arrive the goal, indicating its effectiveness in simple dynamic environments but difficulty in handling cluttered and dynamic scenes. This limitation origins from Panther's detection module, which relies on clustering results to identify dynamic objects. As the environment becomes more complex, clustering requires substantially more time. Furthermore, in intricate environments, increased connections and occlusions between dynamic and static objects degrade classification results.

It is also noteworthy that Chen's method must perform poorly in both scenes if they do not utilize ground truth for perception. This is attributed to their learning-based detection network, which has difficulty detecting objects absent from the training set. In contrast, our system employs the M-detector in its detection module. Although this slightly increases time consumption, it ensures consistent detection results and time consumption across varied environments.

In conclusion, benefiting the M-detector and DynIPC, our system demonstrates superior performance compared to the other two state-of-the-art methods, considering success rate, time consumption, maximum velocity during flight, and adaptability to complex environments.

\subsection{Real-world Experiments}
\begin{figure}[!t]\centering
	\includegraphics[width=4cm]{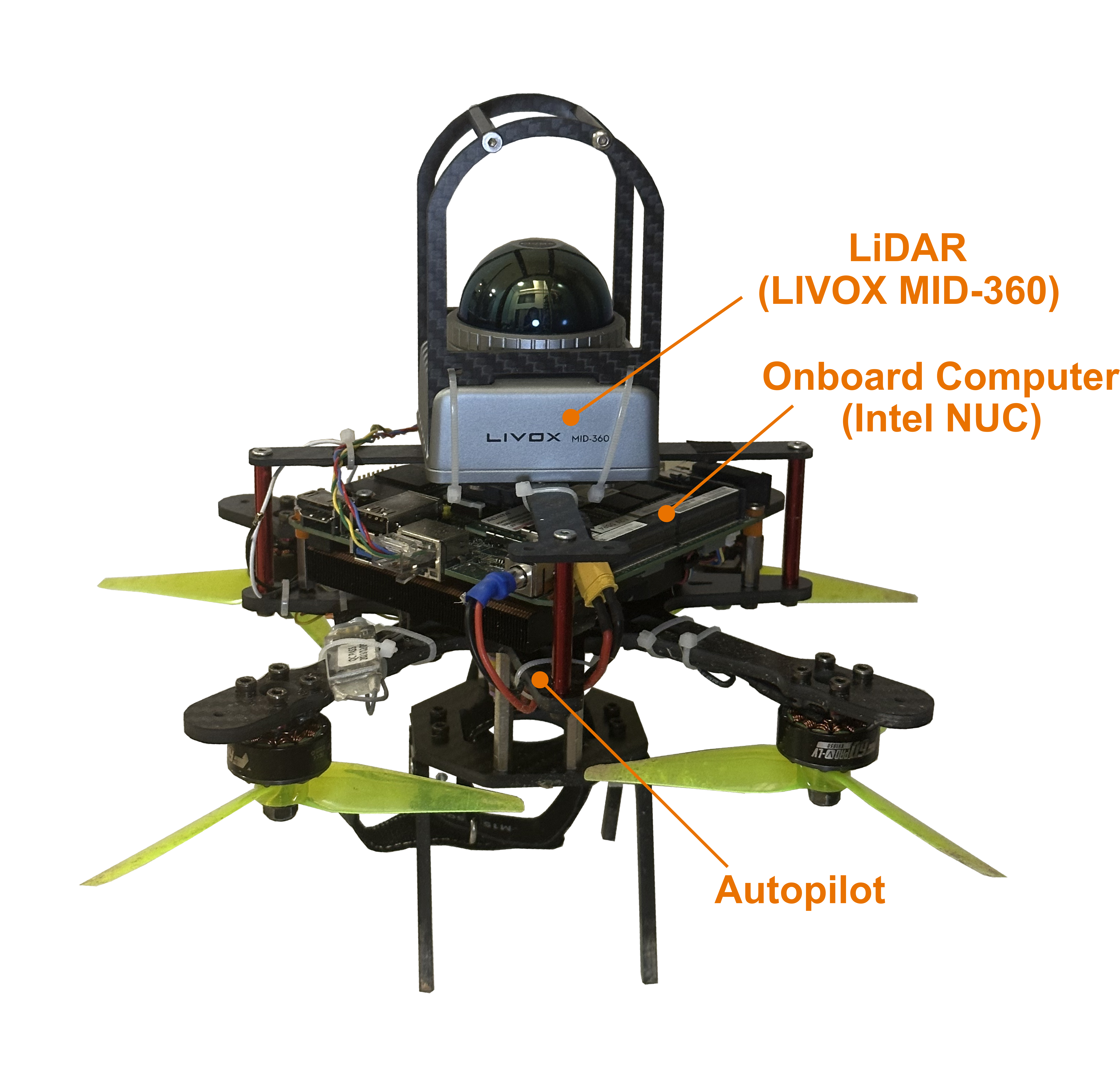}
	\caption{The platform used in the real-world experiments.}\label{fig:uav}
\end{figure}

We also validated the proposed system in real-world scenarios. As shown in Fig. \ref{fig:uav}, the UAV platform primarily comprises a LIVOX Mid-360 LiDAR, an Intel NUC mini-computer with an Intel i7-1260P CPU chip, and an autopilot serving as the controller, which includes a built-in IMU. All computation modules depicted in Fig. \ref{fig:framework}, such as localization, perception, and planning and control, are executed on the onboard computer. For localization, we employed FAST-LIO2 \cite{xu2022fast} to provide odometry, which operates at 100 Hz. The LiDAR collects point cloud data at a frequency of 30 Hz. The back-end processes run at 100 Hz, and its outputs, i.e., control actions, are executed by the autopilot.

The experiments were conducted in a forest with cluttered trees. The UAV was tasked with flying through the forest, avoiding both static and dynamic obstacles, and reaching a goal located 30 meters ahead. Along the UAV's flight path, we set up various dynamic obstacles, including balls of different colors and sizes, hanging pendulums, moving cars, and pedestrians. The detailed parameters for these experiments are listed in Table \ref{tab:param}. The weight matrices in the objective function are diagonal: $\mathbf{R}_p = \mathrm{diag(200, 200, 200)}$, $\mathbf{R}_u =  \mathbf{R}_c  = \mathrm{diag(0, 0, 0)}$, $\mathbf{R}_o =\mathrm{diag(20, 20, 20)}$, and $\mathbf{R}_{v, N} =  \mathbf{R}_{a, N} = \mathrm{diag(30, 30, 30)}$.

\begin{table}[h]
    \centering
    \vspace{-15pt}
    \caption{Parameters of our system.}
    \footnotesize
    \begin{tabular}{c c c}
        \hline
        Parameter & Value & Description \\
        \hline
        $\triangle{t}$ & 0.1 & time step of the MPC \\
        $N$ & 15 & horizon length in the MPC \\
        $d_{rad}$ & 0.3 & quadrotor radius \\
        $k$ & 0.01 & the weight of velocity in inflation distance \\
        $g$ & 9.81 & gravitational acceleration \\
        $v_{i,max}$ & 4 & maximum velocity in the x,y,z direction \\
        $j_{i,max}$ & 20 & maximum jerk in the x,y,z direction \\
        $a_{j,max}$ & $2g$ & maximum acceleration in the x,y direction \\
        $a_{z,min}$ & $-g$ & minimum acceleration in the z direction \\
        $a_{z,max}$ & $2g$ & maximum acceleration in the z direction \\
        $d_{dyn}$ & 0.8 & dynamic obstacles' safety distance \\
        \hline
    \end{tabular}
    \label{tab:param}
\end{table}

\begin{figure}[!t]\centering 
	\includegraphics[width=8cm]{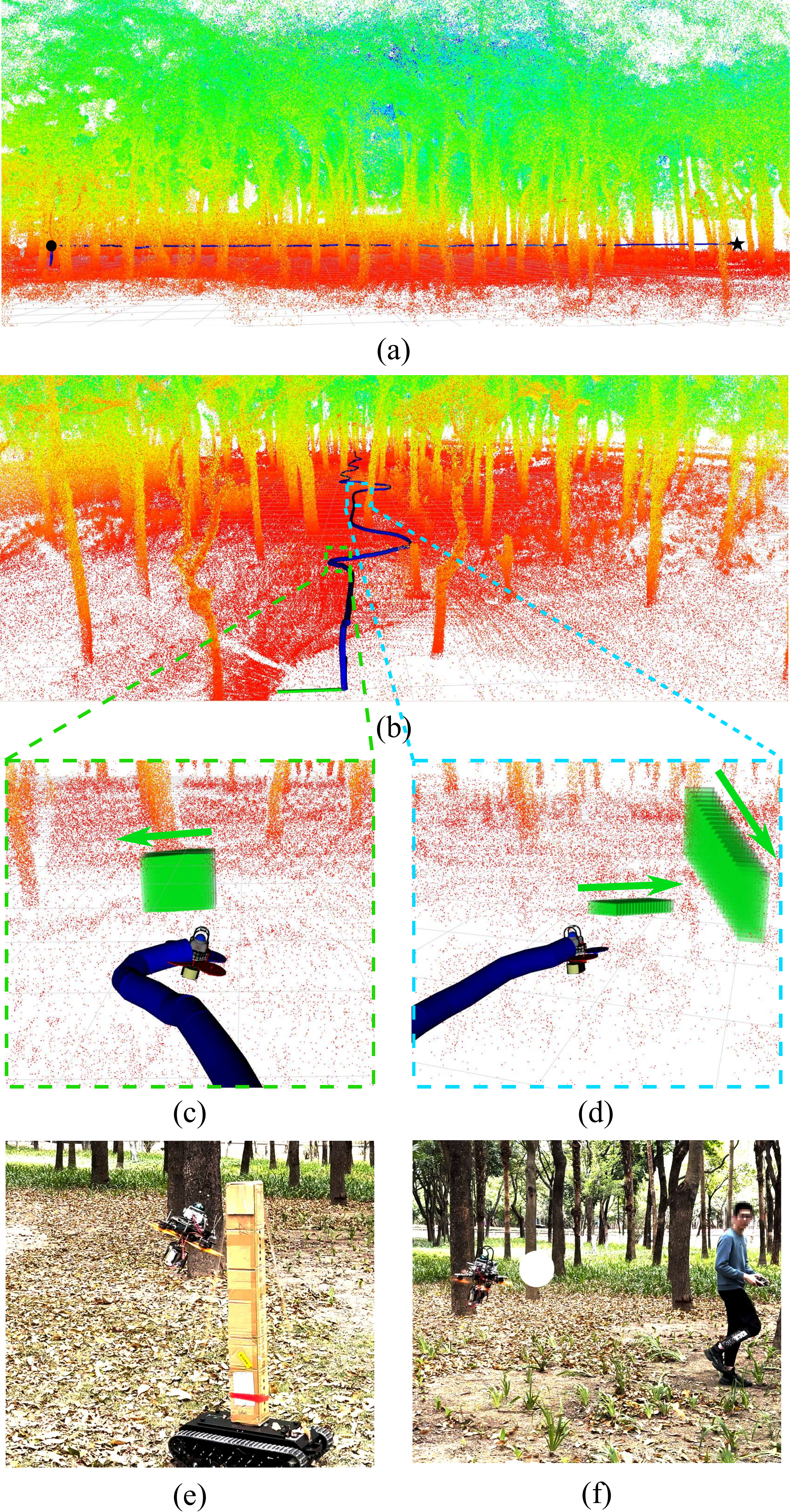}
	\caption{The real-world experiment performed in a cluttered and dynamic forest. (a) The side view of the UAV's trajectory. The dark circle denotes the start point, and the dark pentagram denotes the target. (b) The front view of the UAV's trajectory. (c) The details of the UAV avoiding the moving car. The green boxes denote the dynamic obstacle prediction over the next second. The green arrows denote the movement direction of dynamic obstacles. (d) The details of the UAV avoiding the thrown ball and pedestrian simultaneously. (e) The image corresponding to Fig. \ref{fig:real}(c). (f) The image corresponding to Fig.  \ref{fig:real}(d).}
 \label{fig:real}
\end{figure}

During the experiments, various types, sizes, and numbers of dynamic obstacles appeared in the UAV's Field of View (FoV), and the UAV was required to cope with all these obstacles successfully. Fig. \ref{fig:real} illustrates one such experiment where the UAV successfully avoided both static and dynamic obstacles and reached the target. Two specific avoidance maneuvers along the trajectory are highlighted in Fig. \ref{fig:real}(c) and Fig. \ref{fig:real}(d). As shown in Fig. \ref{fig:real}(c), our system effectively detected and predicted the movement of moving box, triggering the UAV to avoid it. Fig. \ref{fig:real}(d) depicts a scenario where the UAV simultaneously avoided a moving ball and a pedestrian. Additional experiments are referred to our supplementary video by this \href{https://connecthkuhk-my.sharepoint.com/:f:/g/personal/wu2020_connect_hku_hk/Etu5zwJ_GJxAuEx-J8GO354BZRGNPE_dk6qDZBO5DiVpOQ}{\textbf{link}}.

During the flight, the UAV reached a maximum velocity of 4.55 m/s, with the maximum relative speed between the UAV and the moving object at 8.75 m/s. When dynamic obstacles entered the UAV's FoV, the execution latency for perception, replanning, and control was only 11.54 milliseconds. Besides, despite the presence of multiple moving obstacles of varying sizes, colors, and types, our system consistently detected and avoided them, even in complex and cluttered environments. These results demonstrate that our system is capable of navigating through cluttered and dynamic environments with low latency.

\section{Conclusion}
\label{sec:conclusion}
In this paper, we propose a comprehensive LiDAR-based system that includes dynamic environment perception and integrated planning and control for dynamic obstacles. This system enables UAVs to successfully navigate through cluttered and dynamic environments. Multiple experiments, both in simulation and real-world settings, demonstrate that our system can effectively handle fast-moving or sudden-appearing obstacles with high generality in complicated scenes. Furthermore, the entire system can operate efficiently with onboard computation and limited resources with low latency.

\section*{Acknowledgment}
The Authors express their gratitude to Yishen Li for his assistance with the experiments. H.W. Author would like to thank Sile Li and Jinrong Ye for their support in revising the figures for this paper.





\begin{IEEEbiography}[{\includegraphics[width=1in,height=1.25in,clip,keepaspectratio]{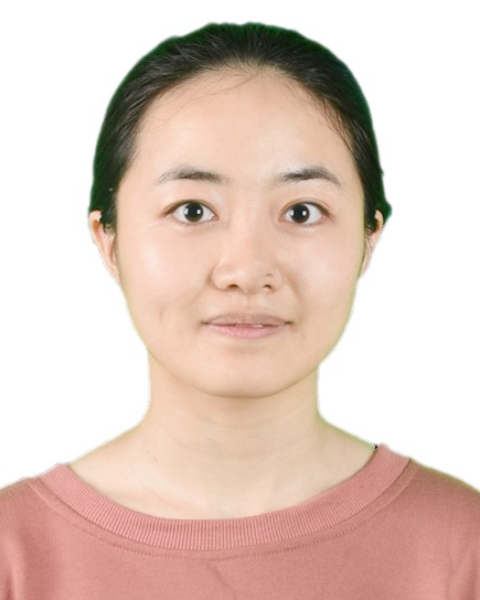}}]
{Huajie Wu} (Graduate Student Member, IEEE) received the B.Eng. degree in Automation from the University of Science and Technology of China (USTC), Hefei, China, in 2020. She is currently working toward the Ph.D. degree in robotics with the Department of Mechanical Engineering, the University of Hong Kong, Hong Kong, China. 

Her research interests include autonomous system, LiDAR perception.
\end{IEEEbiography}

\vspace{-0.5cm}
\begin{IEEEbiography}[{\includegraphics[width=1in,height=1.25in,clip,keepaspectratio]{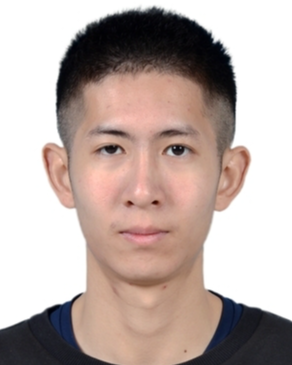}}]
{Wenyi Liu}  (Graduate Student Member, IEEE) received the B.Eng. degree in Communication Engineering from Harbin Institute of Technology (Shenzhen), Shenzhen, China, in 2022. He is currently working toward the PhD degree at the Department of Mechanical Engineering, The University of Hong Kong, Hong Kong. 

His research interests include robot, swarm, planning and control.
\end{IEEEbiography}

\vspace{-0.5cm}
\begin{IEEEbiography}[{\includegraphics[width=1in,height=1.25in,clip,keepaspectratio]{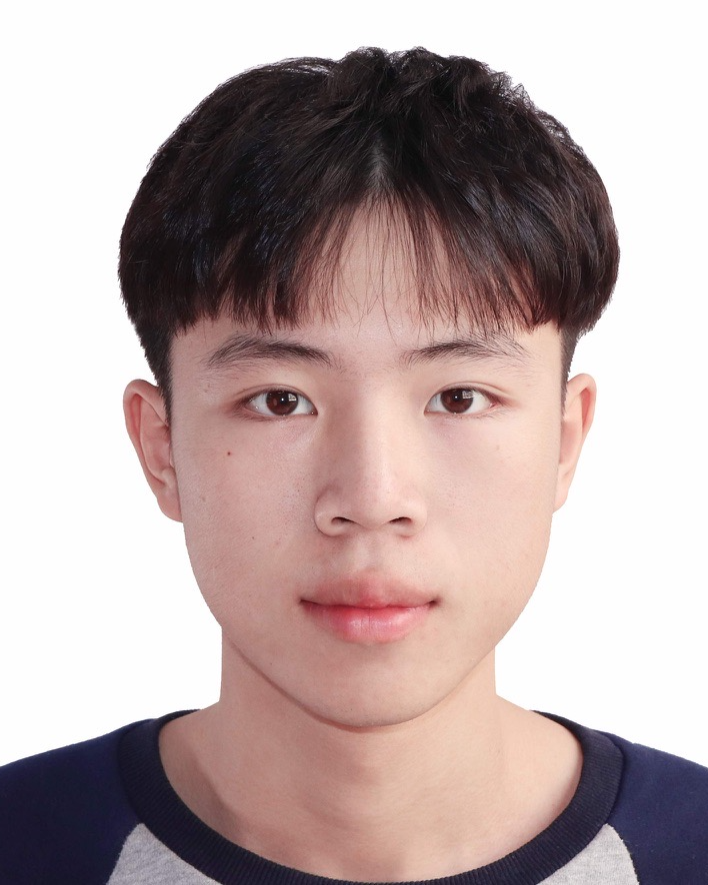}}]
{Yunfan Ren} (Graduate Student Member, IEEE) received the B.Eng. degree in Automation from Harbin Institute of Technology, China, in 2021. He is currently a Ph.D. Candidate with the Department of Mechanical Engineering at The University of Hong Kong, where he is a member of the MaRS Lab. 

His research interests include autonomous navigation, UAV planning, optimal control, and swarm system autonomy.

Mr. Ren was recognized as an Outstanding Navigation Paper Finalist at ICRA 2023 and as both the Best Overall and Best Student Paper Finalist at IROS 2023. He also serves as a reviewer for leading conferences and journals, including IEEE Robotics and Automation Letters (RAL), IROS, and ICRA.
\end{IEEEbiography}

\vspace{-0.5cm}
\begin{IEEEbiography}[{\includegraphics[width=1in,height=1.25in,clip,keepaspectratio]{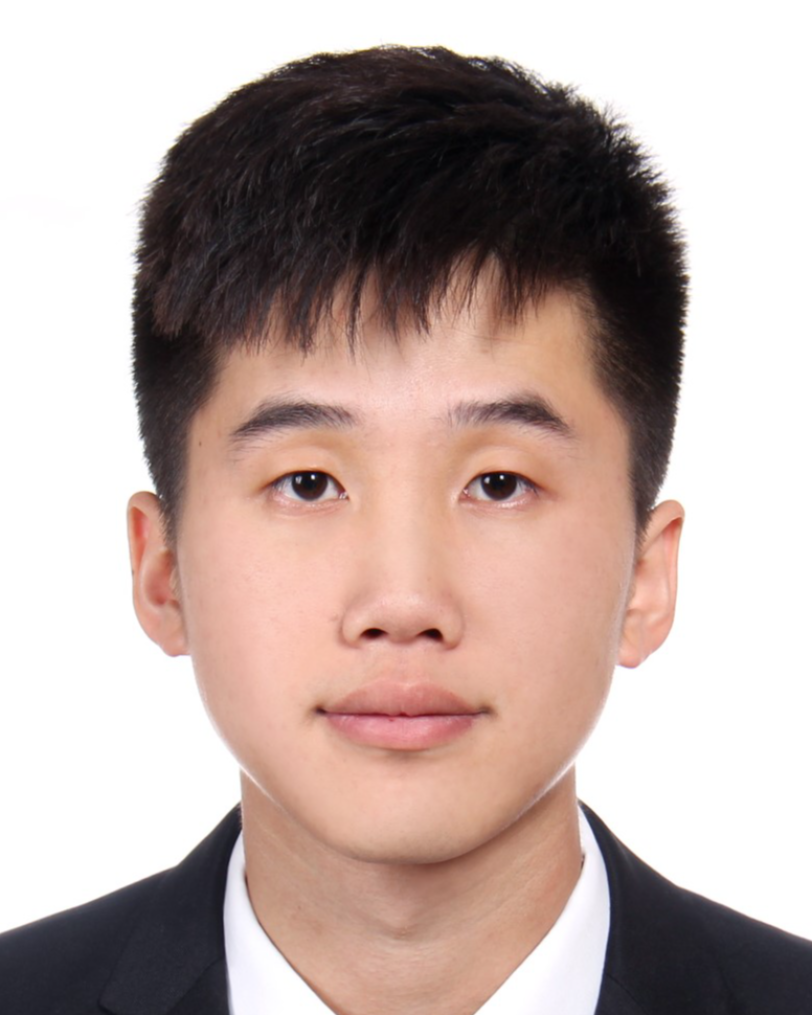}}]
{Zheng Liu} (Graduate Student Member, IEEE) received the B.Eng. degree in automation from the Harbin Institute of Technology, Heilongjiang, China, in 2019 and Ph.D. degree at the University of Hong Kong, Hong Kong, in 2024. 

His research interests include visual or LiDAR-based localization and mapping, sensor fusion and calibration.
\end{IEEEbiography}

\vspace{-0.5cm}
\begin{IEEEbiography}[{\includegraphics[width=1in,height=1.25in,clip,keepaspectratio]{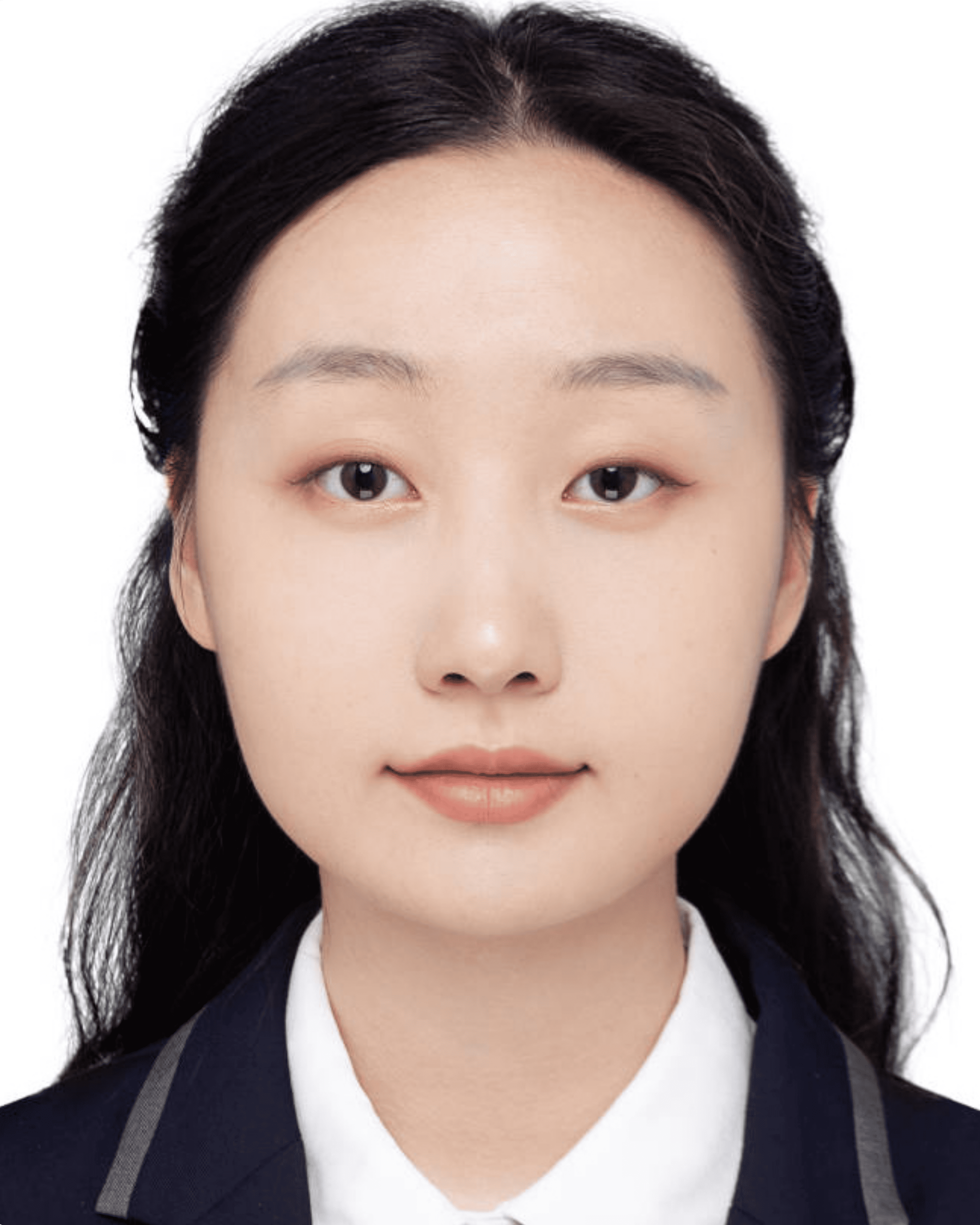}}]
{Hairuo Wei} received the B.Eng. degree in automation, in 2023, from Harbin Institute of Technology, Shenzhen, China. She is currently working toward the MPhil degree in robotics with the University of Hong Kong.

Her current research interests include autonomous aerial robots, swarm robot system and LiDAR mapping.
\end{IEEEbiography}

\vspace{-0.5cm}
\begin{IEEEbiography}[{\includegraphics[width=1in,height=1.25in,clip,keepaspectratio]{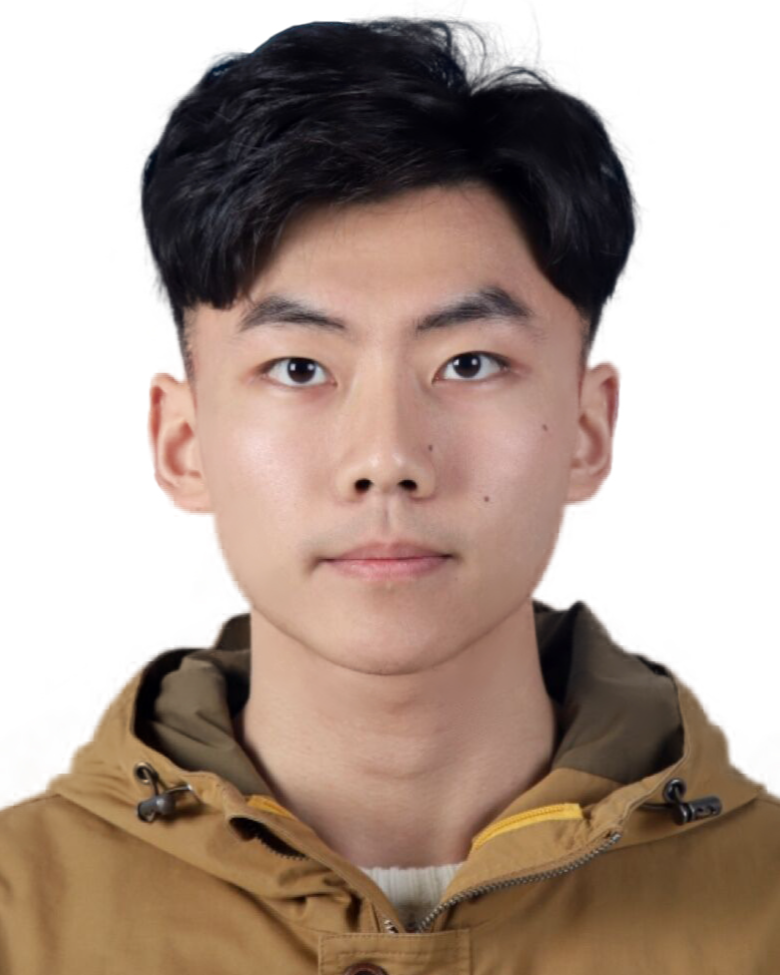}}]
{Fangcheng Zhu} (Graduate Student Member, IEEE) received the B.E. degree in automation in 2021 from the School of Mechanical Engineering and Automation, Harbin Institute of Technology, Shenzhen, China. He is currently working toward the Ph.D. degree in robotics with the Department of Mechanical Engineering, University of Hong Kong (HKU), Hong Kong, China.

His research interests include LiDAR-based simultaneous localization and mapping (SLAM), sensor calibration, and aerial swarm systems.
\end{IEEEbiography}

\vspace{-0.5cm}
\begin{IEEEbiography}[{\includegraphics[width=1in,height=1.25in,clip,keepaspectratio]{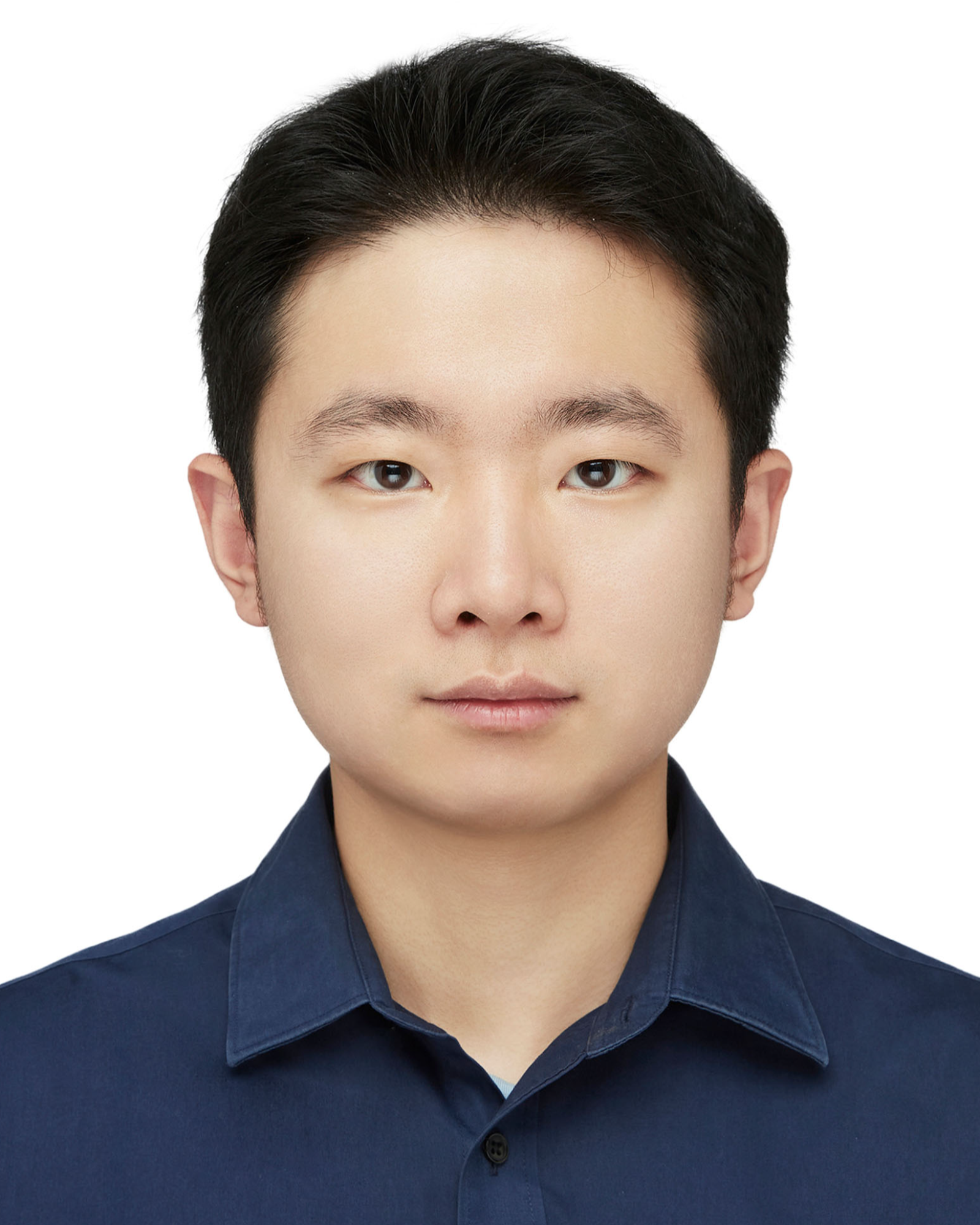}}]
{Haotian Li} (Graduate Student Member, IEEE) received a B.E. degree in Measuring \& Control Technology and Instrumentations from Harbin Engineering University (HEU) in 2021. He is currently a Ph.D. candidate in the Department of Mechanical Engineering, the University of Hong Kong(HKU), Hong Kong, China. 

His current research interests are in robotics, with a focus on unmanned aerial vehicle (UAV) design, and sensor fusion.
\end{IEEEbiography}

\vspace{-0.5cm}
\begin{IEEEbiography}[{\includegraphics[width=1in,height=1.25in,clip,keepaspectratio]{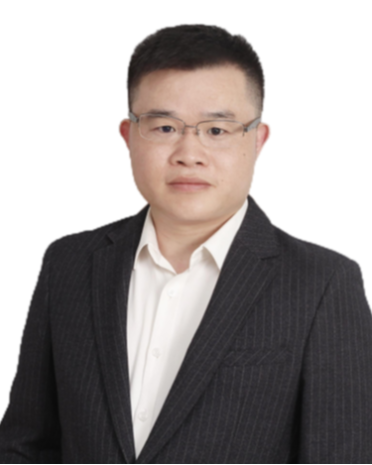}}]
{Fu Zhang} (Member, IEEE) received the B.E. degree in automation from the University of Science and Technology of China (USTC), China, in 2011, and the Ph.D. degree in mechanical engineering from the University of California, Berkeley, Berkeley, CA, USA, in 2015. He joined the University of Hong Kong, Hong
Kong, in 2018, where he is currently an Associate Professor in mechanical engineering. 

His research interests include mechatronics and robotics, with focus on UAV design, control, and Lidar-based navigation.
\end{IEEEbiography}

\end{document}